\documentclass[letterpaper, 10 pt, conference]{ieeeconf}  

\IEEEoverridecommandlockouts                              

\title{\LARGE \bf
Task and Skill Planning:\\
Hierarchical Robot Planning with Black-Box Skills
}

\makeatletter
\let\NAT@parse\undefined
\makeatother

\usepackage{textcomp}
\usepackage[numbers]{natbib}

\usepackage{amsmath,amsfonts,bm}

\def\eqref#1{equation~\ref{#1}}

\def\1{\bm{1}}

\DeclareMathAlphabet{\mathsfit}{\encodingdefault}{\sfdefault}{m}{sl}
\SetMathAlphabet{\mathsfit}{bold}{\encodingdefault}{\sfdefault}{bx}{n}

\def\gA{{\mathcal{A}}}

\def\gC{{\mathcal{C}}}

\def\gG{{\mathcal{G}}}

\def\gI{{\mathcal{I}}}

\def\gM{{\mathcal{M}}}

\def\gO{{\mathcal{O}}}

\def\gR{{\mathcal{R}}}

\def\gU{{\mathcal{U}}}
\def\gV{{\mathcal{V}}}
\def\gW{{\mathcal{W}}}
\def\gX{{\mathcal{X}}}

\usepackage{stmaryrd}
\usepackage{graphicx}
\usepackage{amssymb}

\usepackage{subfigure}
\usepackage{hyperref}
\hypersetup{
    colorlinks=true,
    allcolors=black,
}

\newtheorem{definition}{Definition}

\author{Benned Hedegaard$^{1\dagger*}$, Yichen Wei$^{1\dagger}$, Ziyi Yang$^1$, Ahmed Jaafar$^1$, \\
Stefanie Tellex$^1$, George Konidaris$^1$, and Naman Shah$^{1,2{\S}}$%
\thanks{$^{1}$Department of Computer Science, Brown University, Providence, RI. $^{2}$Allen Institute for Artificial Intelligence, Seattle, WA. $^{\S}$Work primarily completed while at Brown University.}%
\thanks{$^{\dagger}$Equal contribution}%
\thanks{$^{*}$Corresponding author: benned\_hedegaard@brown.edu}%
}

\begin{document}

\maketitle
\thispagestyle{empty}
\pagestyle{empty}

\begin{abstract}
Task and motion planning (TAMP) is a well-established approach for solving long-horizon robot planning problems.
Although TAMP methods have historically assumed that each task-level robot action, or skill, can be reduced to kinematic motion planning, recent work has explored integrating closed-loop controllers and learned skills into TAMP-style systems.
Our approach integrates pre-existing, heterogeneous robot skills---including learned, force-controlled, and black-box policies---into a hierarchical planner while preserving the object-centric failure reasoning of typical TAMP solvers.
We leverage Composable Interaction Primitives (CIPs) to synthesize head and tail motion plans bridging consecutive skills, facilitating both planning-time refinement and execution-time adjustment.
We validate our \emph{Task and Skill Planning (TASP)} approach through real-world experiments on a bimanual manipulator and a mobile manipulator,
demonstrating that CIPs enable diverse robots to combine heterogeneous skills to solve complex, long-horizon tasks, including multi-room mobile manipulation problems with non-monotonic task structure.
\end{abstract}
\section{Introduction} \label{sec:intro}

Task and motion planning (TAMP) combines discrete task-level reasoning with continuous motion planning to solve long-horizon robot planning problems~\citep{shah2020anytime, garrett2021integrated}.
Recent work has extended TAMP-style methods beyond purely kinematic actions by incorporating learned skills~\citep{wang.garrett.ea2021,silver.athalye.ea2023,OptimisticRL2024}, closed-loop controllers~\citep{TAMPER2024,curtis2024partially}, and uncertainty-aware execution~\citep{curtis2024partially,garrett.paxton.ea2020}.
However, these methods typically assume a particular policy structure or controller class for the integrated skills.
Therefore, the remaining challenge is not whether skills can be used within hierarchical planning, but how to plan effectively when a robot is equipped with a heterogeneous, pre-existing inventory of general-purpose skills.

In practice, robot skills often span multiple implementation classes, including learned policies, force-controlled behaviors, built-in robot skills, and trajectory playback routines.
Skill controllers may be closed-loop, stochastic, or otherwise opaque to the planner, and may not permit a uniform internal model or parameterization.
Nevertheless, a planner must reason about when such skills can be executed and how their effects may enable or obstruct future actions.
This is especially important for non-downward-refinable actions~\citep{marthiAngelicSemanticsHighLevel2007}, where resolving failure may require reasoning about object obstructions and counterfactual scene modifications.

We address this setting by exposing planner-facing geometric structure for otherwise black-box skills. We use a \emph{kinematic envelope} to characterize where a skill can be initiated and its nominal geometric effects, allowing us to package skills as Composable Interaction Primitives~\cite{abbatematteo2024composable} that synthesize motion plans between consecutive skills.
We instantiate this idea as \emph{Task and Skill Planning (TASP)}, a hierarchical planning framework that retains motion-planning-based feasibility checking and object-centric geometric reasoning without requiring explicit internal models of skill controllers.
We evaluate our approach on two real-world robots, a bimanual manipulator and a mobile manipulator, demonstrating that TASP can plan using motion-planned, learned, force-controlled, and built-in robot skills to solve long-horizon real-world tasks,
including mobile manipulation problems with non-monotonic task structure.




\begin{figure}[t!]
\centering
\subfigure[\texttt{Pick(?eraser)}]{\includegraphics[width=0.49\columnwidth]{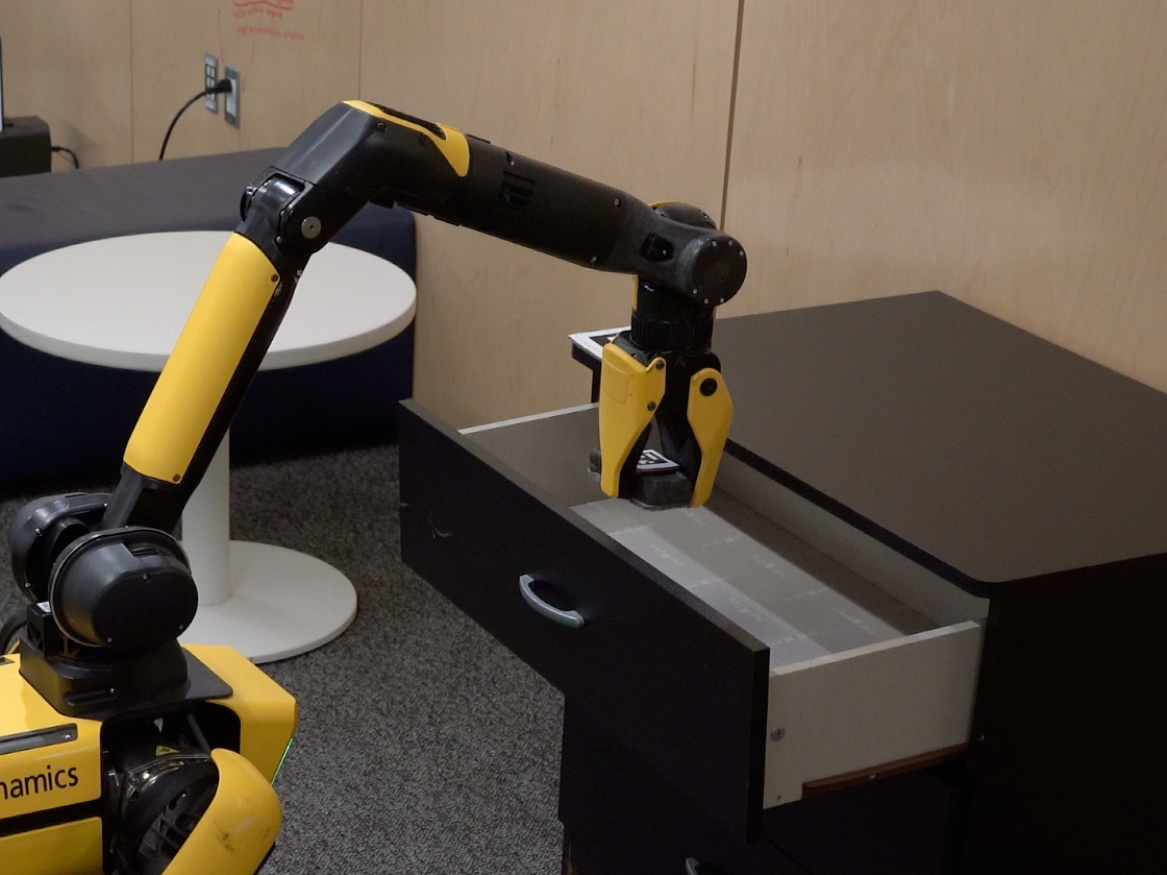}\label{fig:1a-pick}}
\subfigure[\texttt{Erase(?board)}]{\includegraphics[width=0.49\columnwidth]{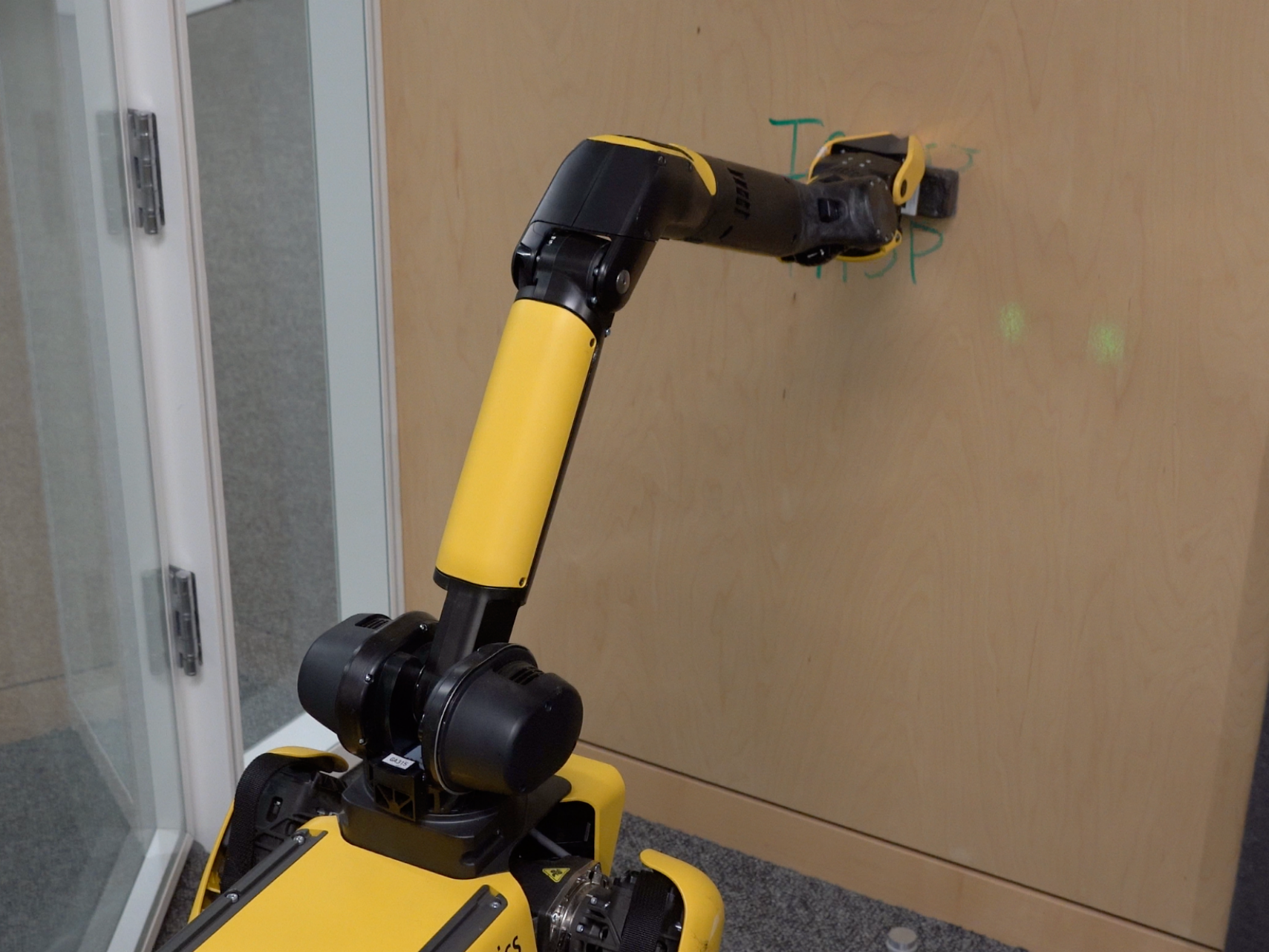}\label{fig:1b}} \\[-0.4em]

\subfigure[\texttt{Grasp(?jar)}]{\includegraphics[width=0.49\columnwidth]{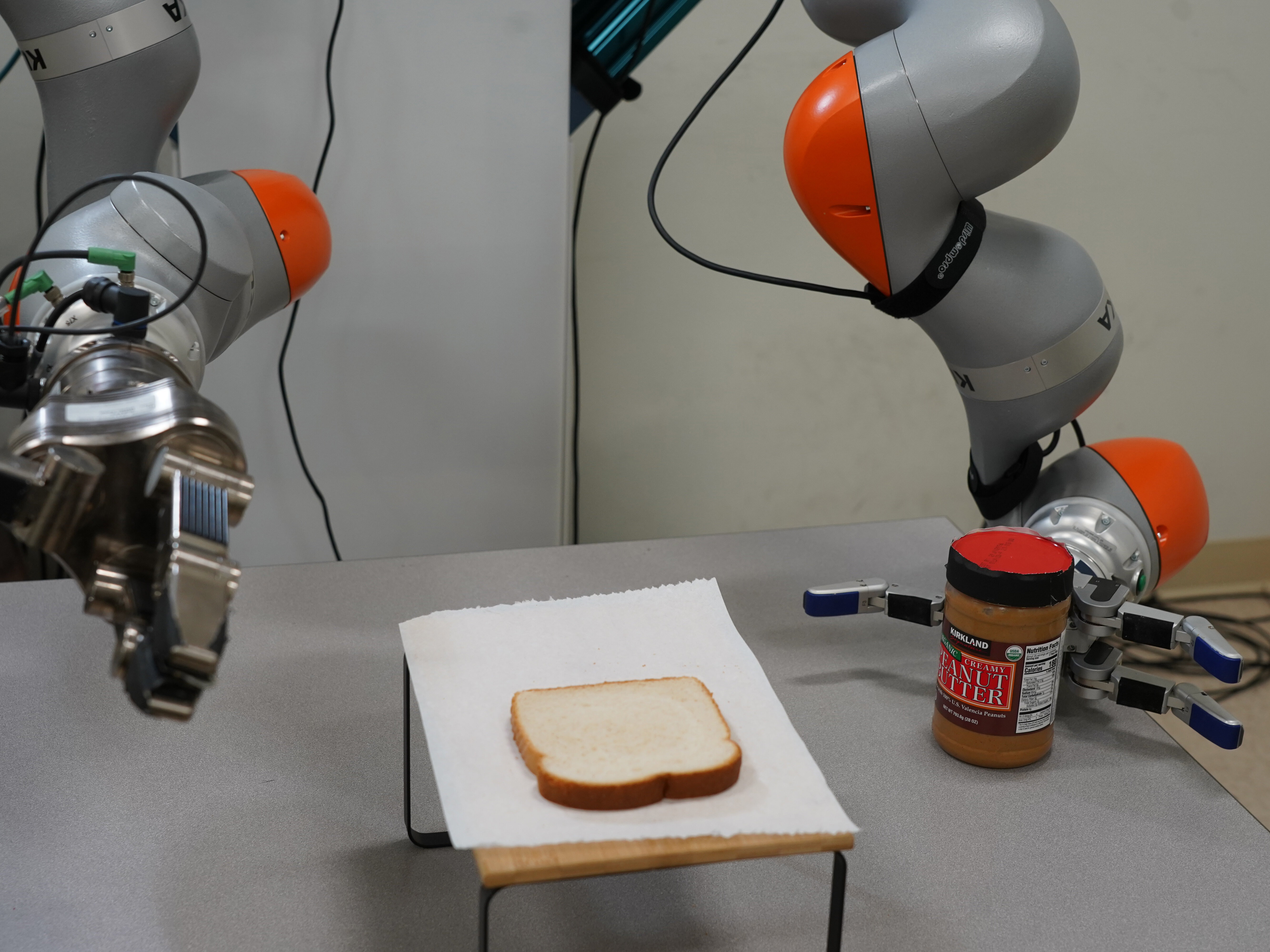}\label{fig:1c}}
\subfigure[\texttt{Spread(?knife,} \texttt{?bread)}]{\includegraphics[width=0.49\columnwidth]{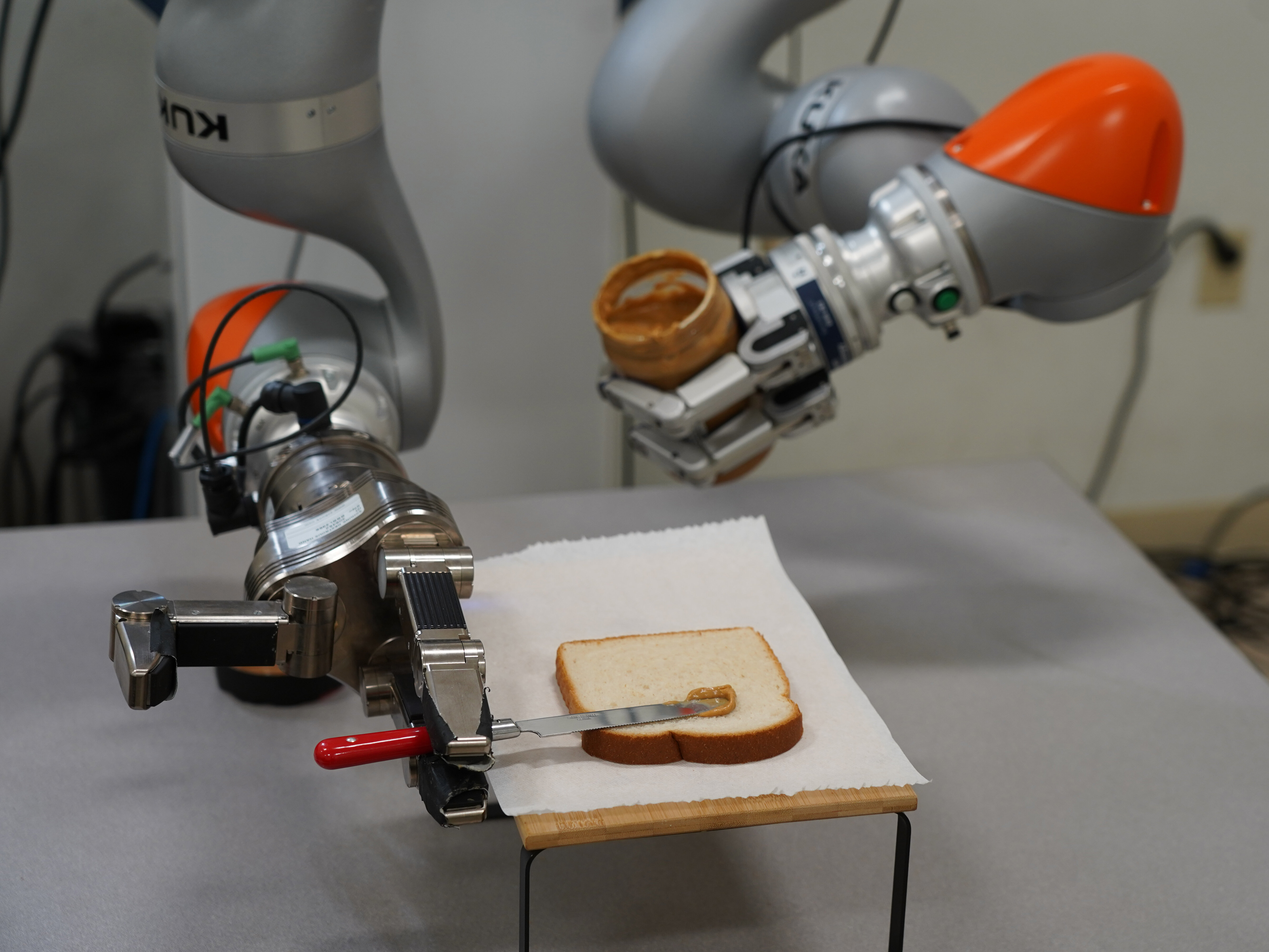}\label{fig:1d}}
\label{fig:intro_fig}
\caption{A Boston Dynamics Spot and a bimanual manipulator use task and skill planning (TASP) to solve long-horizon hybrid robot planning problems. On the left, the robots execute motion-planning-based skills to pick up an eraser (Fig.~\ref{fig:1a-pick}) and a jar of peanut butter (Fig.~\ref{fig:1c}). On the right, the robots use general-purpose skills to maintain force-controlled contact while erasing a whiteboard (Fig.~\ref{fig:1b}) and spreading peanut butter (Fig.~\ref{fig:1d}).}
\end{figure}




\section{Preliminaries}
\label{sec:prelim}

This paper brings together ideas from different domains of robotics, including motion planning, symbolic reasoning, task and motion planning, and skill learning. This section outlines the foundational concepts necessary to understand the proposed methodology.

\subsection{Motion Planning}
A robot state is represented by a \emph{configuration}, i.e., an assignment of values to all robot joints.
A motion plan moves the robot from one collision-free configuration to another.
Formally, let 
$\gC = \gC_\mathrm{free} \cup \gC_\mathrm{obs}$ denote the robot's configuration space~\citep{lavalle2006planning}, where $\gC_\mathrm{free}$ is the set of collision-free configurations and $\gC_\mathrm{obs}$ is the set of configurations in collision with an obstacle.
Let $c_i \in \gC_\mathrm{free}$ and $c_g \in \gC_\mathrm{free}$ be the initial and goal configurations, respectively.

\begin{definition}
    A \textbf{motion planning problem} is a $4$-tuple $\langle \gC, F, c_i, c_g \rangle$, where $\gC$ is the configuration space, $F: \gC \rightarrow \{0,1\}$ is a collision indicator such that $F(c) = 1$ iff $c \in \gC_\mathrm{obs}$, and $c_i$ and $c_g$ are the initial and goal configurations.
\end{definition}

A solution to a motion planning problem is a collision-free trajectory $\tau: [0,1] \rightarrow \gC$ such that $\tau(0) = c_i$, $\tau(1) = c_g$, and $F(\tau(t)) = 0$ for all $t \in [0,1]$.
%
%
Although motion planning is a fundamental component of robot control, it is insufficient for very long-horizon problems due to its continuous state space and large branching factor.
Moreover, it does not capture the perceptual reasoning or closed-loop feedback required by many robot skills.

\subsection{Symbolic Task Planning} \label{subsec:symbolic-planning}
Symbolic task planning is widely used for long-horizon reasoning because it operates over a discrete state space with a restricted branching factor.
Planning domains and problems are often represented in relational PDDL~\citep{McDermott_1998_PDDL}.
Formally, a PDDL domain is a tuple $\langle \gV, \gA \rangle$, where $\gV$ is a vocabulary of parameterized predicates and $\gA$ is a set of high-level actions defined over that vocabulary.
A high-level state is a set of true grounded predicates, i.e., predicates whose parameters are bound to concrete objects.
Each action $a \in \gA$ is defined as a tuple $\langle \Theta_a, \textsc{Pre}_a, \textsc{Eff}_a \rangle$, where $\Theta_a$ is the set of action parameters, $\textsc{Pre}_a$ is the set of preconditions, and $\textsc{Eff}_a$ is the set of effects.

A symbolic planning problem is a tuple $\langle \gU, s_i, \gG \rangle$, where $\gU$ is the universe of objects, $s_i$ is the initial symbolic state,
and $\gG$ is a set of goal conditions specified as grounded predicates.
A solution to a symbolic planning problem is a sequence of actions that transforms $s_i$ into a state that satisfies $\gG$.
Although symbolic task planning is well-suited to long-horizon reasoning, its plans are not directly executable on a robot and must be refined into motion plans or other low-level controllers.






\subsection{Task and Motion Planning}
Task and motion planning (TAMP) refers to a class of methods~\citep{srivastava2014combined,shah2020anytime,dantam2018incremental,garrett2021integrated} that interleave high-level symbolic task planning with low-level motion planning to solve complex robot planning problems.
TAMP approaches search for a high-level plan such that each symbolic action can be refined into a feasible motion plan for real-world execution.
Following~\citet{shah2020anytime}, we define a task and motion planning problem as follows: 
\begin{definition}
    \label{def:tamp}
   A \textbf{task and motion planning problem} is a tuple $\langle \gW, \alpha, \gV_\alpha, \gA_\alpha, \gU, w_i, \gW_g, \gG, \Gamma \rangle$.
   Here, $\gW$ is the system configuration space of the robot and environment; $\alpha$ is an abstraction function; $\gV_\alpha$ is a set of symbolic relations defined using $\alpha$; $\gA_\alpha$ is a set of high-level robot actions, where each action $a \in \gA_\alpha$ is executed via motion planning; $\gU$ is a universe of objects; $w_i \in \gW$ is the initial system configuration, inducing the initial symbolic state $s_i = \alpha(w_i)$; $\gW_g \subseteq \gW$ is a set of goal configurations; $\gG$ is the corresponding set of high-level goal conditions; and $\Gamma$ is the inverse abstraction function used to generate motion plans for symbolic actions.
\end{definition}

A solution to a task and motion planning problem is a task-level plan with corresponding motion plans that, when executed from $w_i$, reach a configuration in $\gW_g$.
This formulation is sufficient when every high-level action can be refined into motion planning.
However, some robot actions involve non-kinematic constraints, such as sustained contact or closed-loop feedback, and therefore cannot be captured by motion planning alone.
The next section extends this formulation to such hybrid settings.


\section{Hybrid Robot Planning with General-Purpose Policies}
\label{sec:formal}

The task and motion planning formulation in Def.~\ref{def:tamp} focuses on kinematic properties of the robot and environment, such as configurations and poses.
Many real-world tasks require additional forms of control, including compliant control, sustained contact, or feedback-driven execution.
These tasks may also modify non-spatial properties of objects, such as whether a surface is wet or a dish is dirty.


\paragraph*{Environment Model}

We model the environment as a collection of objects and robots.
Let $\gO$ denote the set of objects and $\gR$ the set of robots. 
For each object, a spatial function $P^o: \gO \rightarrow SE(3)$ maps the object to its 6-DoF pose in a fixed reference frame.
%
%
We also define object-specific observation functions $\Phi = \{\Phi_o\}_{o\in\gO}$, where each $\Phi_o$ is a set of attribute classifiers $\{\phi_i^o\}$ for object $o$, each classifying a distinct attribute such as color, temperature, or cleanliness.
%

Each robot is modeled as a kinematic tree of rigid-body links connected by joints.
A configuration function $C: \gR \rightarrow \gC$ maps each robot $r \in \gR$ to its current configuration, and a spatial function $P^r: \gR \rightarrow SE(3)$ maps each robot to the pose of its base link.
%
%
An attachment function $\zeta: \gR \rightarrow 2^{\gO}$ maps a robot to the set of objects currently attached to it.

We use first-order logic to represent the symbolic state of the environment.
The universe is $\gU = \gO \cup \gR$, and the vocabulary $\gV$ includes the spatial functions $P^o$, $P^r$, and $C$, the attachment function $\zeta$, and the observation functions $\Phi$.
The high-level state space is defined as the set of grounded predicate assignments over $\gU$ and $\gV$.

\paragraph*{General-Purpose Robot Skills}

We assume that each robot is equipped with object-centric skills that modify the state of the environment.
Let $\gA$ denote the set of such skills.
Each skill $a^r \in \gA$ for robot $r \in \gR$ is modeled as a parameterized option
$\langle \Theta_a, \gI_a, \beta_a, \pi_a \rangle$,
where $\Theta_a \subseteq \gO$ is the set of object arguments of the skill, $\gI_a$ is an initiation condition, $\beta_a$ is a termination condition, and $\pi_a$ is the skill policy.
Both $\gI_a$ and $\beta_a$ are first-order logic formulas over the vocabulary $\gV$.
A skill can be executed from any state $x \models \gI_a$ and follows policy $\pi_a$ until reaching a state $x' \models \beta_a$, where the skill terminates.


Some skills can be implemented using motion planning because they exploit the structure of the robot configuration space,
whereas others (e.g., sustained contact behaviors) require control strategies not easily expressed in kinematic terms.
These non-kinematic skills may be implemented using techniques such as behavior cloning~\citep{padalkar2023open} or reinforcement learning~\citep{haramatientity}. We now define a hybrid robot planning problem to model such skills.
\begin{definition}
    \label{def:hybrid_robot_planning_problem}
    A \textbf{hybrid robot planning problem} is a tuple $\gM = \langle \gU, \gV, \gX, \gA, x_i, \gX_g \rangle$, where $\gU = \gO \cup \gR$ is the universe, $\gV$ is the vocabulary, $\gX$ is the state space, $\gA$ is a set of robot skills, $x_i \in \gX$ is the initial state, and $\gX_g \subseteq \gX$ is the set of goal states.
\end{definition} 

A solution to a hybrid robot planning problem is a sequence of skills $[a_1, \ldots, a_n]$ that, when executed from $x_i$, reaches a state $x_n \in \gX_g$.
In principle, such problems could be solved by model-based search if the initiation and termination sets of all skills were explicitly available.
In our setting, however, these sets are induced by continuous geometric and controller-level constraints, and may depend on scene context not explicitly captured by the symbolic state.
As a result, even when the symbolic actions predict that two skills $a_i$ and $a_j$ are compatible, there may not exist states $x,x' \in \gX$ such that $x \models \beta_{a_i}$, $x' \models \gI_{a_j}$, and the robot can transition between $x$ and $x'$ through collision-free motion planning or skill execution.

To support planning in this setting, each skill $a \in \gA$ must expose three properties to the planner:
$(i)$ the robot must be able to reach a state $x \in \gI_{a}$ through collision-free motion;
$(ii)$ after executing the skill, the robot must be able to return from a state $x \in \beta_{a}$ to free space; and
$(iii)$ the planner must be able to identify compatible initiation and termination states $x_i, x_j \in \gX$ such that $x_i \models \gI_{a}$, $x_j \models \beta_{a}$, and executing $\pi_{a}$ from $x_i$ can realize the transition from $x_i$ to $x_j$.
These requirements motivate the planning abstractions introduced in the next section, which enable structured motion planning to connect and compose otherwise black-box robot skills.




\section{Hierarchical Planning for Hybrid Robot Planning Problems}
\label{sec:approach}

We now present \emph{Task and Skill Planning (TASP)}, our approach for solving hybrid robot planning problems. TASP is designed to satisfy the three planner-facing properties above, enabling the composition of general-purpose object-centric robot skills. We first discuss our solution to the first two properties (Sec.~\ref{subsec:cips}) that make individual skills composable, and then describe how our solution to the third property (Sec.~\ref{subsec:tasp}) facilitates composing skills using TAMP-style hierarchical planning.

\subsection{General-Purpose Object-Centric Robot Skills as CIPs}
\label{subsec:cips}

\begin{figure*}[t]
\subfigure[]{\includegraphics[width=0.24\textwidth]{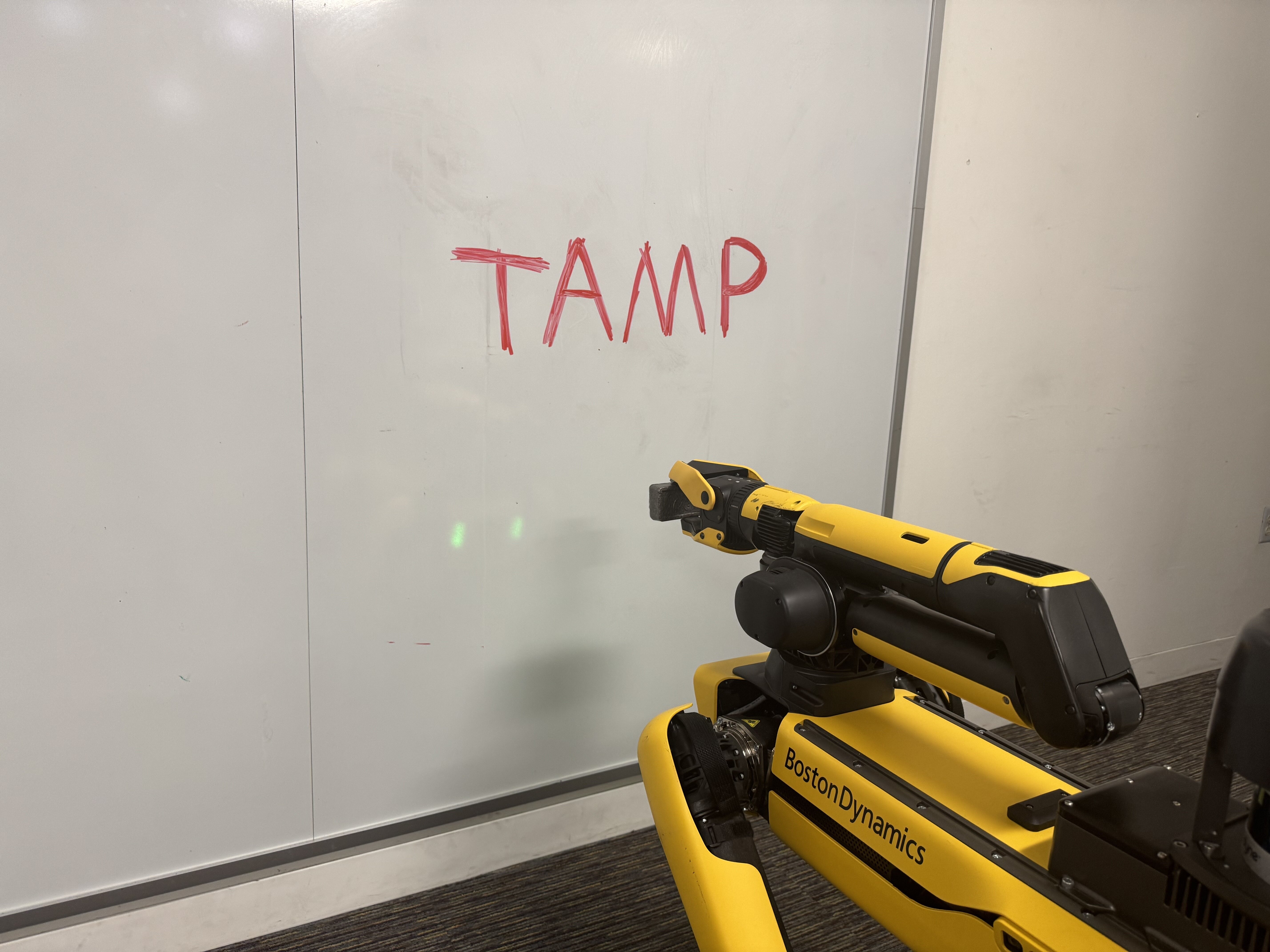}\label{fig:erase-1}}\hfill
\subfigure[]{\includegraphics[width=0.24\textwidth]{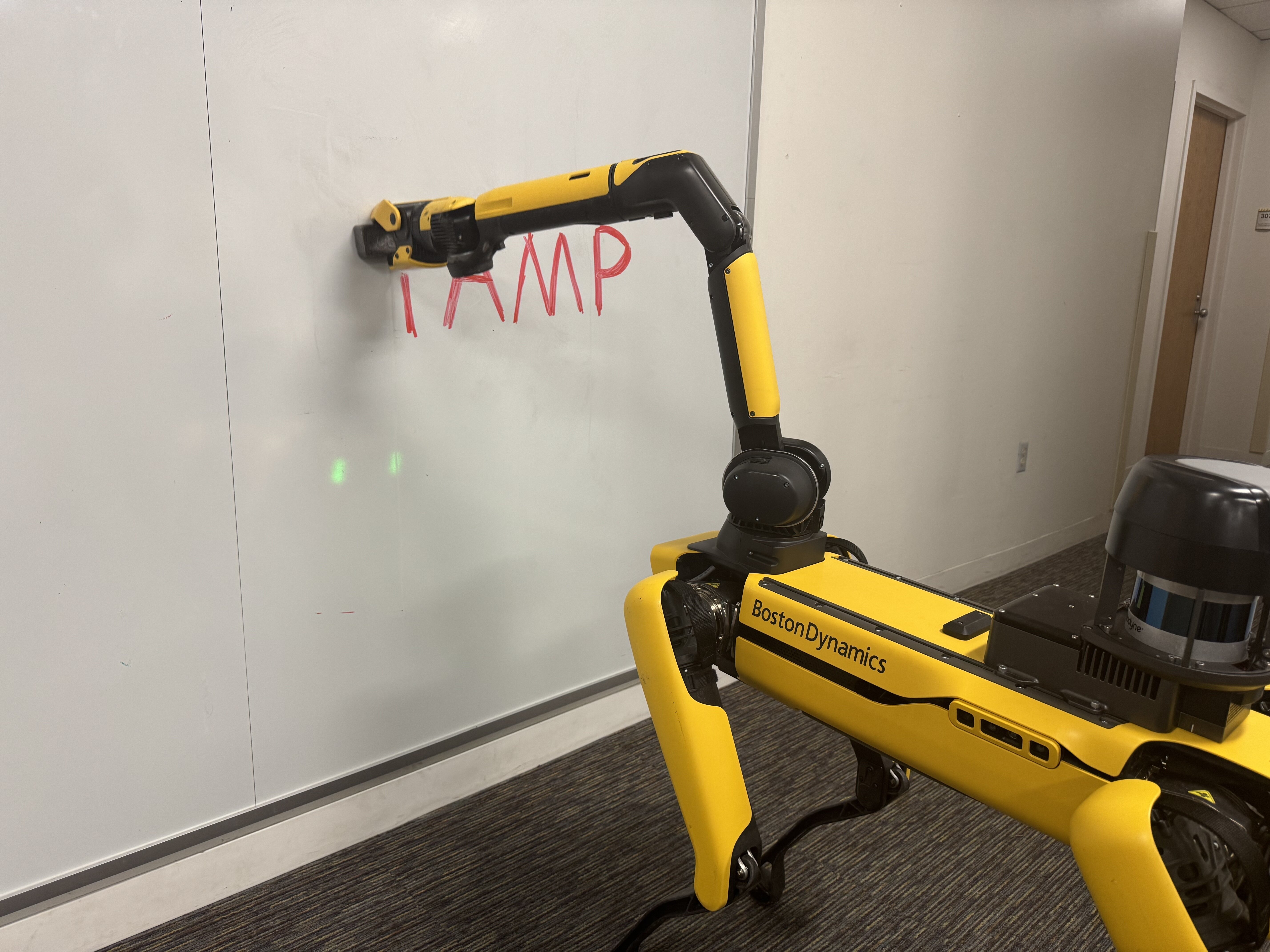}\label{fig:erase-2}}\hfill
\subfigure[]{\includegraphics[width=0.24\textwidth]{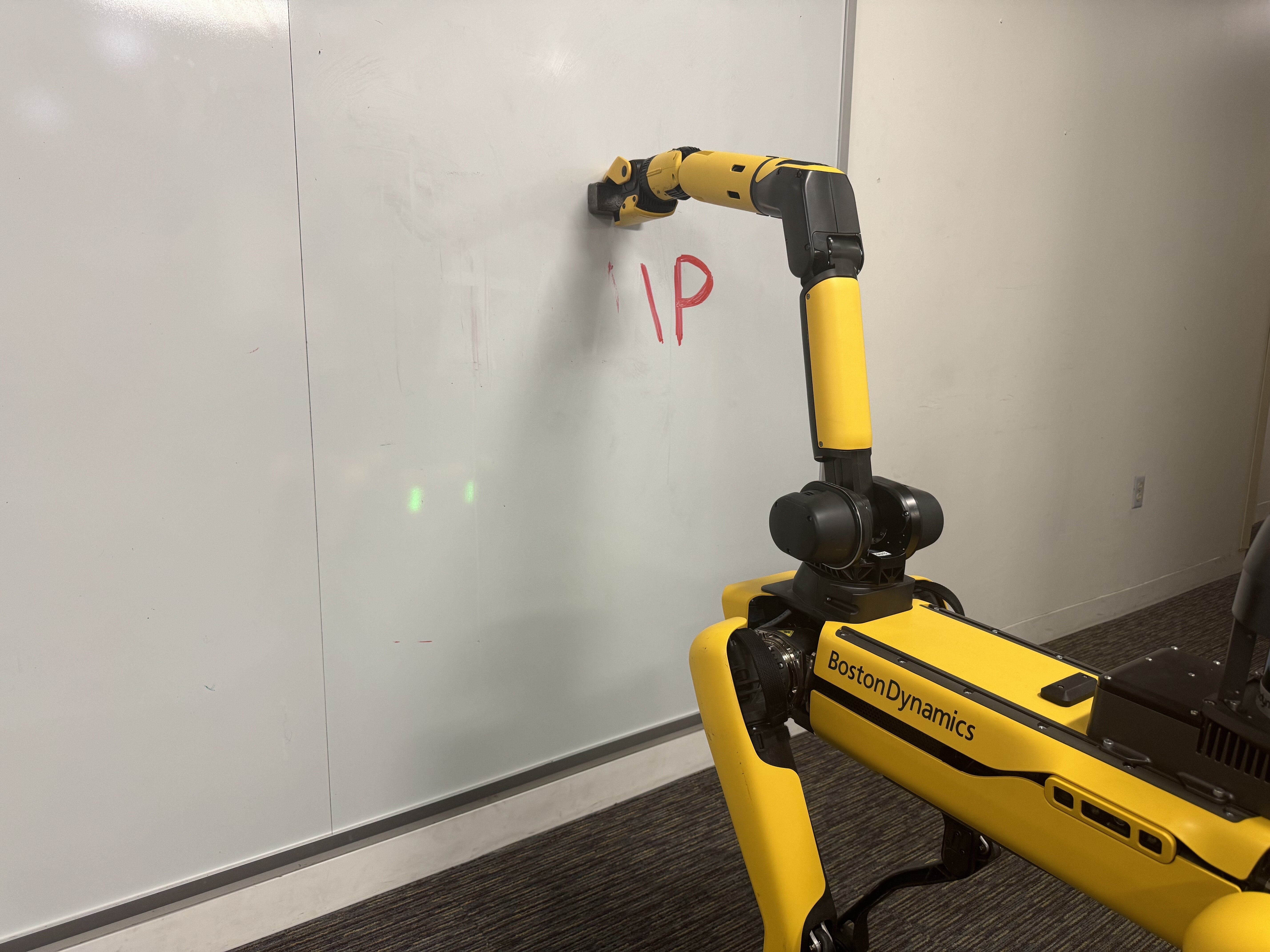}\label{fig:erase-3}}\hfill
\subfigure[]{\includegraphics[width=0.24\textwidth]{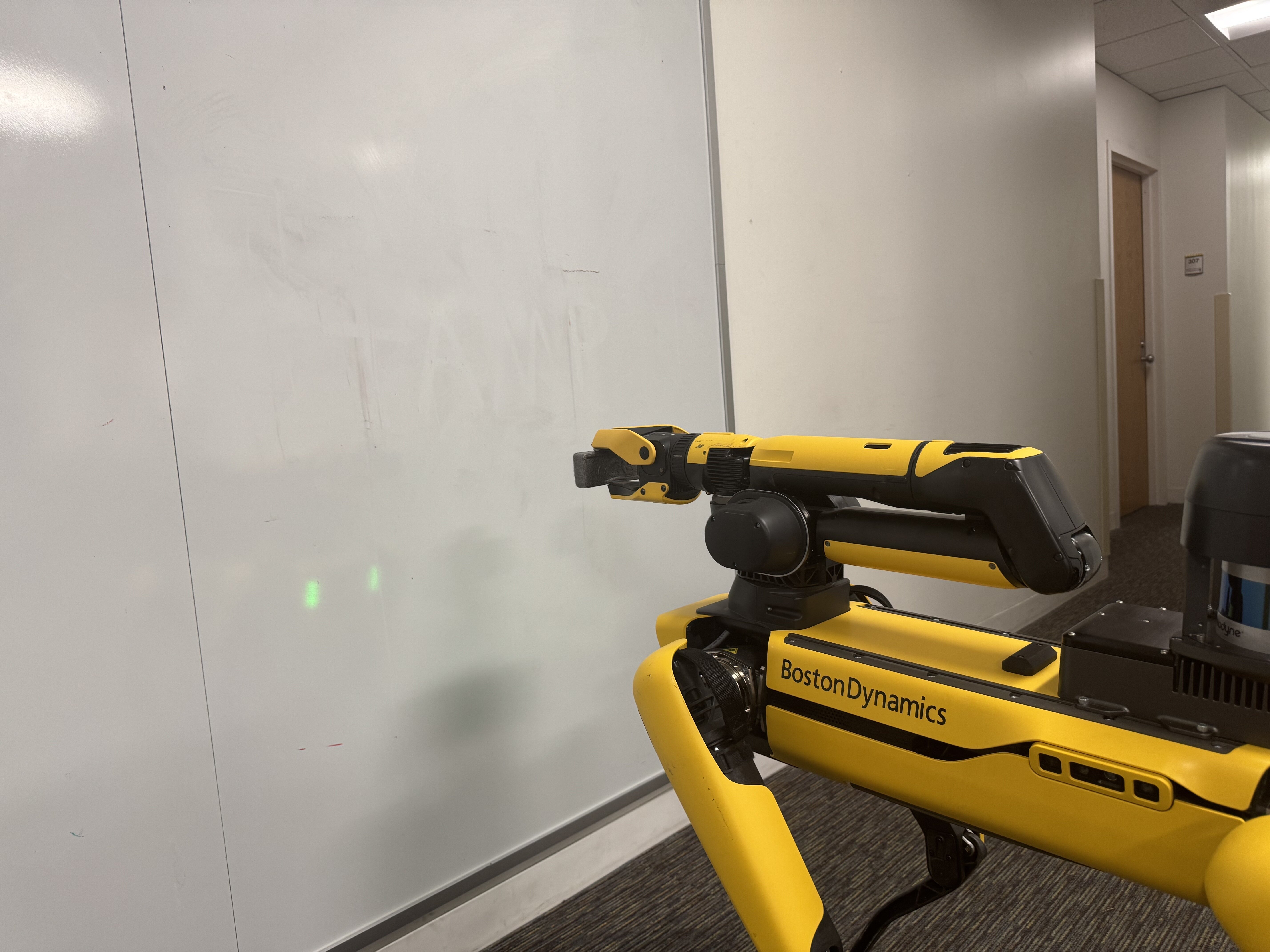}\label{fig:erase-4}}
\caption{Example keyframes showing the environment state before, during, and after the \texttt{Erase(?board)} force-controlled skill.
In Fig.~\ref{fig:erase-1}, the Spot robot can reach the whiteboard, but its gripper is stowed. Therefore, the robot is not in a configuration that fulfills the initiation condition for the \texttt{Erase(?board)} skill.
Fig.~\ref{fig:erase-2} shows the state after Spot has executed a \emph{head motion plan} to reach a configuration that fulfills the initiation condition for the \texttt{Erase(?board)} skill.
Fig.~\ref{fig:erase-3} depicts a state during the execution of the \texttt{Erase(?board)} skill.
Finally, Fig.~\ref{fig:erase-4} shows the state after Spot has executed a \emph{tail motion plan} to stow its arm, resulting in a configuration that is no longer in the termination set of the \texttt{Erase(?board)} skill.}
\label{fig:erase}
\end{figure*}

The core properties needed to enable the composition of general-purpose object-centric robot skills are the robot's ability to transition into and out of the skills' initiation and termination sets. \citet{abbatematteo2024composable} develop a key policy class named \emph{Composable Interaction Primitives (CIPs)} that wraps free-space motion plans around object-centric skills. These motion plans allow the robot to use the structure of the configuration space to transition to configurations that satisfy the initiation and termination conditions of its skills.

Intuitively, a CIP is composed of three components: a motion plan that takes the robot to the initiation set of the object-centric skill, the object-centric skill policy, and a motion plan that takes the robot out of the termination set of the skill. The two free-space motion planning components provide a natural means to compose different skills, enabling the robot to transition between the initiation and termination sets of separate skills without requiring them to intersect.

Consider the example of the Spot robot that erases a whiteboard in Fig.~\ref{fig:erase}. Here, the initial configuration in Fig.~\ref{fig:erase-1} does not satisfy the initiation condition of the \texttt{Erase(?board)} skill. However, the robot can use a motion plan to move from its current configuration to the configuration shown in Fig.~\ref{fig:erase-2}, which is in the initiation set of the skill. The robot can then use the pre-trained policy to erase the whiteboard, as shown in Fig.~\ref{fig:erase-3}. Lastly, the Spot uses a motion plan to reach a state, shown in Fig.~\ref{fig:erase-4}, that does not satisfy the termination conditions of the \texttt{Erase(?board)} skill. From there, Spot can again use a motion plan to reach the initiation set of the next skill.

Formally, let $a \in \gA$ be a robot skill operating on object(s) $o_1,\dots,o_k$, with initiation and termination conditions $\gI_a(P^r, P^{o_{1:k}}, C, \Phi_{o_{1:k}})$ and $\beta_a(P^r, P^{o_{1:k}}, C, \Phi_{o_{1:k}})$, respectively. Here, $\gI_a(P^r, P^{o_{1:k}}, C, \Phi_{o_{1:k}})$ expresses that the initiation condition for the robot skill $a$ is a formula over spatial pose functions $P^r$ and $P^{o_{1:k}}$, the configuration function $C$, and object-specific observation functions $\Phi_{o_{1:k}}$. For example, for the skill \texttt{Erase(?board)}, the initiation condition is a function of the robot pose, the robot configuration, the pose of the whiteboard, and whether the whiteboard is dirty.

We define a kinematic projection $\lambda$
that abstracts away object-specific observation functions from a skill's initiation condition. Specifically, the projection $\lambda$ maps the initiation condition $\gI_a(P^r, P^{o_{1:k}}, C, \Phi_{o_{1:k}})$  to its \emph{kinematic envelope} $\gI_a(P^r, P^{o_{1:k}}, C)$. The kinematic envelopes of initiation and termination conditions are only defined over spatial functions, and hence, can be achieved using kinematic motion planning. We formally define a composable interaction primitive for an object-centric robot skill as follows.
\begin{definition}
Let $a = \langle \Theta_a, \gI_a, \beta_a, \pi_a \rangle$ be an object-centric robot skill with initiation condition $\gI_a$ and termination condition $\beta_a$.
A \textbf{composable interaction primitive} (CIP) $\tilde{a}$ for the robot skill $a$ is defined as a tuple $\tilde{a} = \langle \Theta_{a},  \gI_a, \beta_a, \pi_a, h, t, \lambda  \rangle$.
Here, the parameters $\Theta_a$, initiation condition $\gI_a$, termination condition $\beta_{a}$, and policy $\pi_a$ are identical to the robot skill $a$.
$\lambda$ is the kinematic projection; $h$ is the head motion plan from a configuration $x \not\models \gI_a$ to a configuration $x \models \lambda(\gI_a)$ and $t$ is the tail motion plan from a configuration $x \models \beta_a$ to a configuration $x \not\models \lambda(\beta_{a})$. 
\end{definition}

Using CIPs, we can compose individual skills using motion planning. We next describe how TASP uses these abstractions within a hierarchical planner to identify long-horizon plans.



\subsection{Hierarchically Composing CIPs}
\label{subsec:tasp}

In this section, we describe our approach to efficiently composing CIPs into long-horizon plans.
We also discuss how we compute endpoints for head and tail motion plans, enabling skill composition and successful skill execution.

Let $\tilde{a}_i$ and $\tilde{a}_j$ be two CIPs for the object-centric robot skills $a_i$ and $a_j$, respectively. The major challenges in composing the skills $a_i$ and $a_j$ are: $(i)$  we must ensure that the head motion plan $h_{\tilde{a}_i}$ reaches a configuration $x_n \models \lambda(\gI_{a_i})$ such that $\pi_{a_i}$ can be successfully executed given the other objects in the environment, $(ii)$ we must ensure that the tail motion plan $t_{\tilde{a}_i}$ ends at a configuration $x_m$ from which the next head motion plan $h_{\tilde{a}_j}$ can be executed, and $(iii)$ we must select the configurations $x_n$ and $x_m$ such that there exist collision-free head and tail motion plans.
One could attempt to identify such motion plans using sampling-based motion planners such as the RRT~\citep{lavalle2001rapidly} or PRM~\citep{kavraki1996probabilistic}.
However, this is poorly suited to the structure of our problem: as in multi-modal motion planning, feasible solutions may depend on transitions through narrow skill-conditioned compatibility regions in the continuous state space~\cite{hauser.latombe2010}.
We therefore cast CIP composition as a hierarchical TAMP problem.

\begin{figure*}[t!]
    \centering

    \subfigure[\texttt{Grasp(j0)}]{%
        \includegraphics[width=0.28\linewidth]{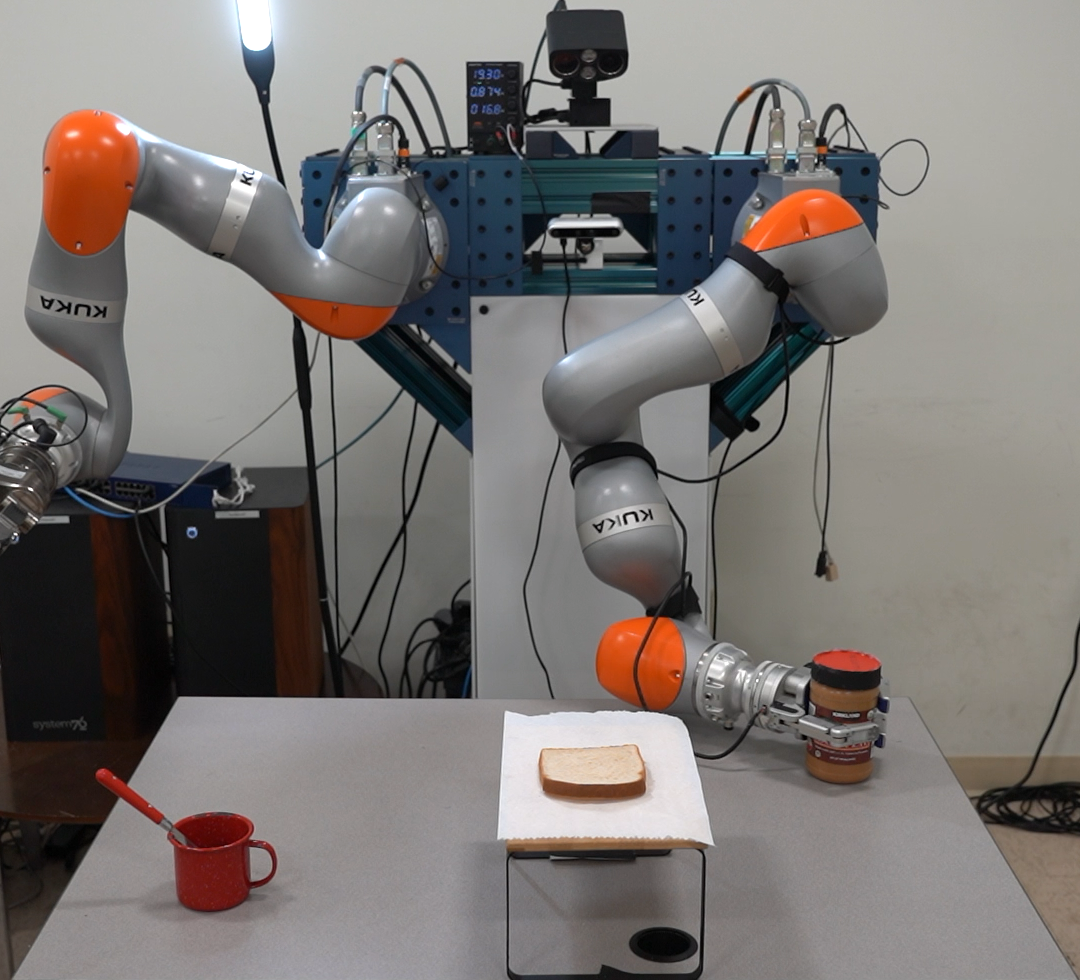}
        \label{fig:dorfl-grasp-pb}} \hspace{1.5em}
    \subfigure[\texttt{Open(j0)}]{%
        \includegraphics[width=0.28\linewidth]{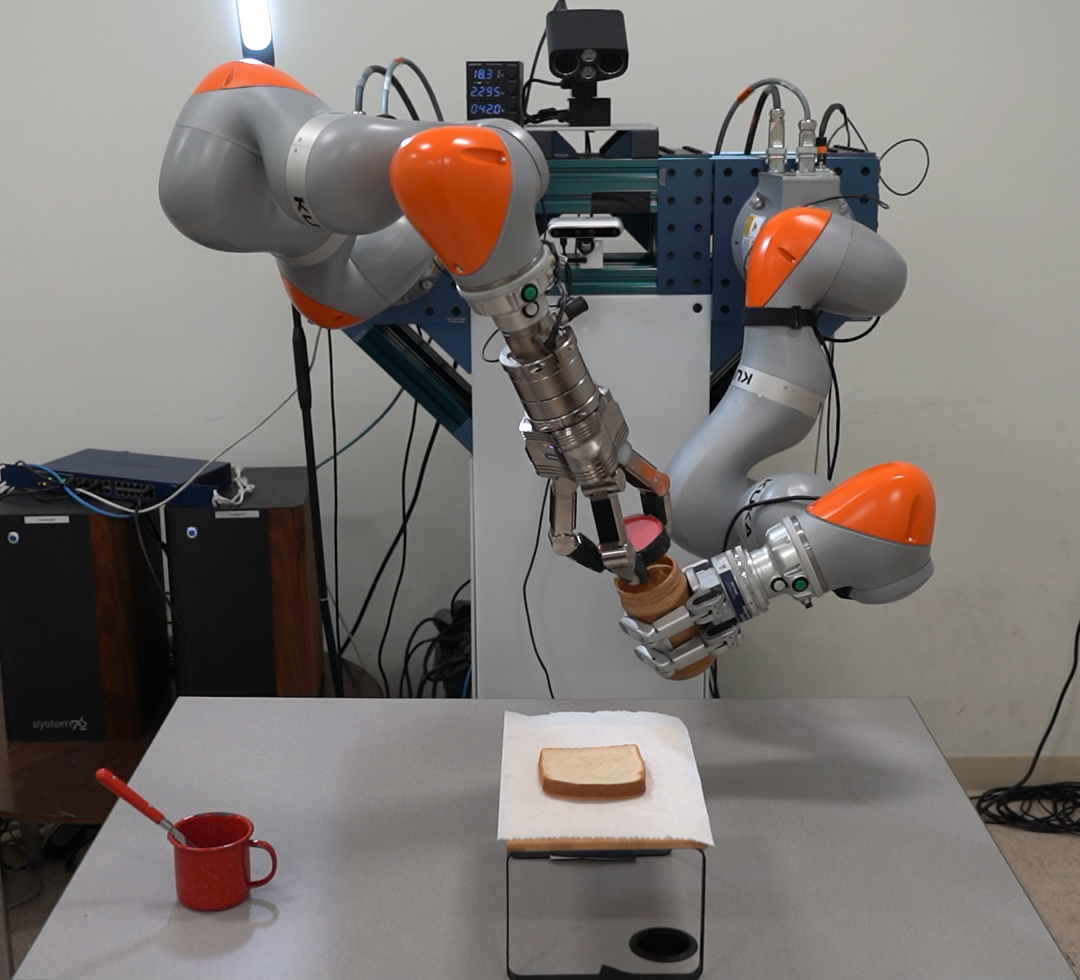}
        \label{fig:dorfl-open-pb}} \hspace{1.5em}
    \subfigure[\texttt{Grasp(k0)}]{%
        \includegraphics[width=0.28\linewidth]{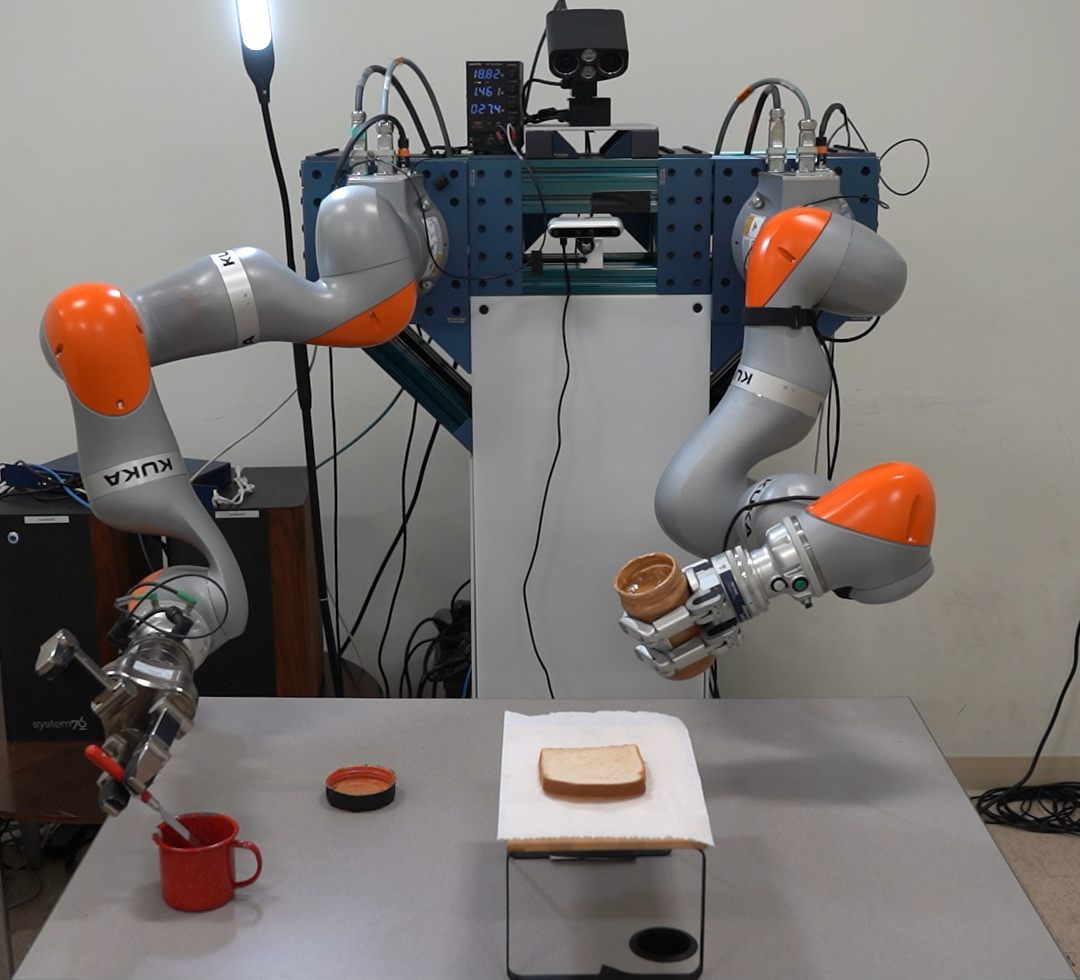}
        \label{fig:dorfl-grasp-knife}} \vspace{0.25em}

    \subfigure[\texttt{Scoop(k0,} \texttt{j0)}]{%
        \includegraphics[width=0.28\linewidth]{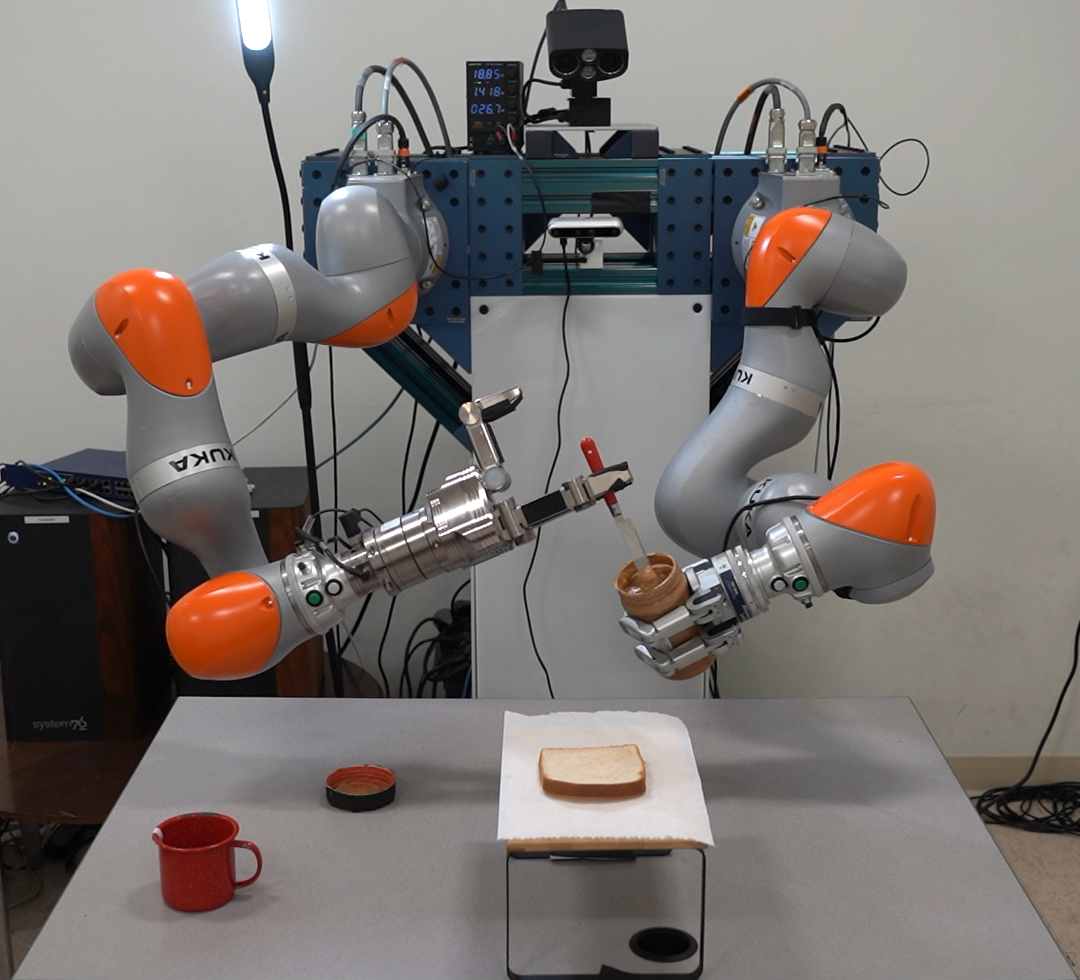}
        \label{fig:dorfl-scoop}} \hspace{1.5em}
    \subfigure[\texttt{Spread(k0,} \texttt{b0)}]{%
        \includegraphics[width=0.28\linewidth]{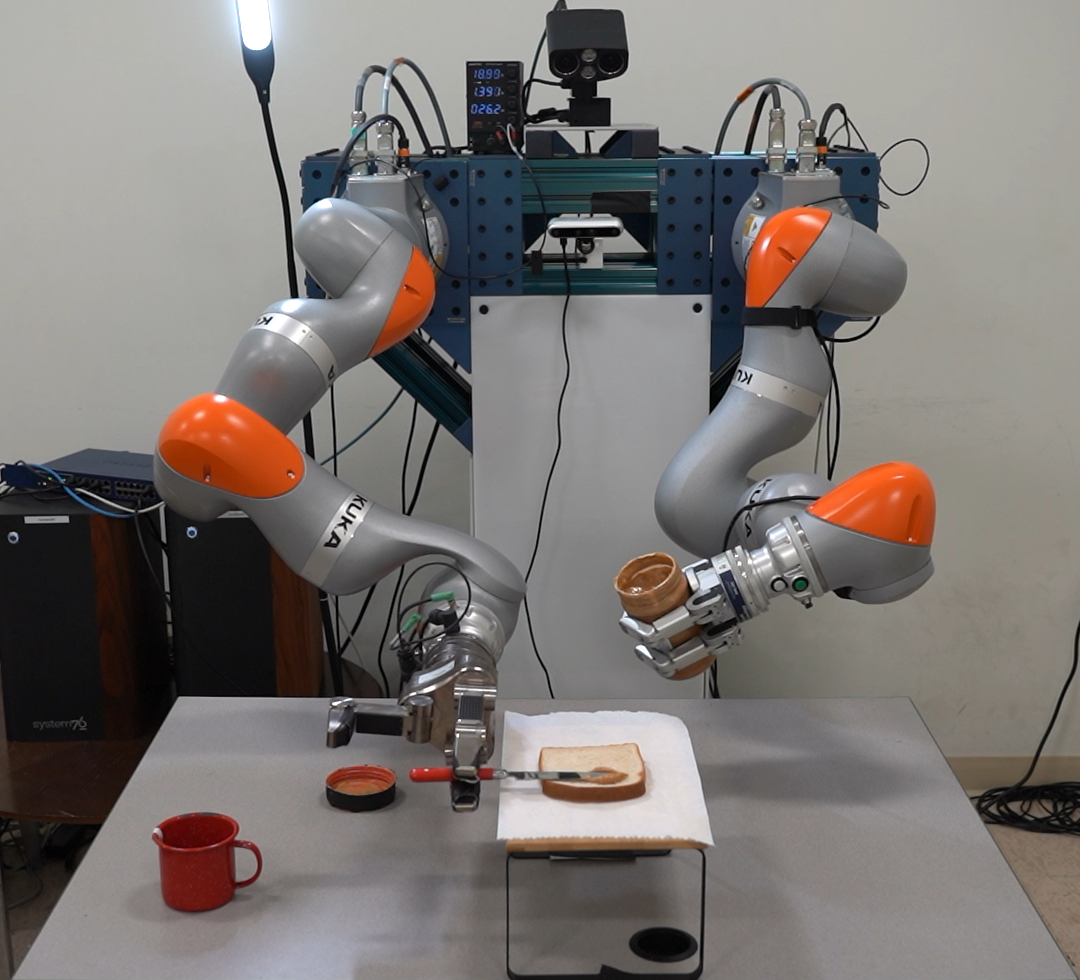}
        \label{fig:dorfl-spread}}
    \\

    \caption{We conduct real-world experiments on two robot platforms to demonstrate the versatility of our task and skill planning approach.
    In our bimanual manipulation setting, the robot can pick up a jar and a kitchen knife using a \texttt{Grasp(?object)} skill implemented using motion planning (Figs.~\ref{fig:dorfl-grasp-pb} and \ref{fig:dorfl-grasp-knife}).
    The bimanual manipulator has three additional skills that combine trajectory playback and impedance control:
    the robot uses \texttt{Open(?jar)} to open a jar of peanut butter using both grippers (Fig.~\ref{fig:dorfl-open-pb});
    scoops peanut butter onto a knife using \texttt{Scoop(?knife,} \texttt{?jar)} (Fig.~\ref{fig:dorfl-scoop}); and
    spreads peanut butter by executing \texttt{Spread(?knife,} \texttt{?bread)} (Fig.~\ref{fig:dorfl-spread}).}
    \label{fig:dorfl_execution}
\end{figure*}

Specifically, we use the ATAM algorithm~\citep{shah2020anytime} and entity abstraction~\citep{shah2020anytime} to reduce the problem of efficiently combining CIPs to a TAMP problem.
Entity abstraction converts each CIP into a high-level action with discrete parameters, including symbols for the end configurations, the head and tail motion plans, and the skill policy.

We modify the ATAM approach to use black-box skill policies instead of calling a motion planner while refining the high-level action arguments. ATAM uses a high-level symbolic planner to compute a high-level plan, applies the inverse abstraction function $\Gamma$ to reduce each high-level action into a motion planning problem, and uses a sampling-based motion planner to compute motion plans for these refinements. It performs backtracking-based search to recompute a high-level plan until it finds one with valid low-level refinements for each of its high-level actions. This allows the ATAM algorithm to efficiently sample endpoints for head and tail motion plans, overcoming the aforementioned challenges.



\section{Real-World Experiments}
\label{sec:eval}

We validate our method on two robot platforms: a KUKA iiwa bimanual setup and a Boston Dynamics Spot. We first describe the experimental setup for each platform (Sec.~\ref{subsec:experimental-setup}) and then discuss our results (Sec. \ref{subsec:results}). Videos, code, and supplementary material are available at \url{https://benned-h.github.io/tasp/}.

\subsection{Experimental Setup} \label{subsec:experimental-setup}

We now describe the experimental setup of our real-world evaluation, beginning with the shared problem setting and then specifying the platform-specific hardware, robot skill inventories, and perception pipelines.

\paragraph*{Problem Setting}

In each experiment, our Task and Skill Planning (TASP) framework is given as input a PDDL-style symbolic planning problem specifying $\langle \mathcal{U}, s_i, \mathcal{G} \rangle$, which are the universe, initial symbolic state, and goal conditions (see Sec. \ref{subsec:symbolic-planning}). The initial low-level state is modeled as a kinematic tree comprised of known object poses and collision geometry.
Given this information, our method must plan a feasible sequence of skills that can be executed on the robot to satisfy the goal conditions $\mathcal{G}$.

\subsubsection{Bimanual Manipulator Setting}
\label{subsubsec:dorfl-experimental-setup}

\paragraph*{Robot Hardware}

The bimanual manipulator uses two KUKA LBR iiwa 7 R800 manipulators, one equipped with a BarrettHand BH8-282 gripper, and the other with a SCHUNK Dextrous Hand 2.0 gripper.
The robot collects RGB-D data using an Intel RealSense D455 depth camera.

\paragraph*{Skill Inventory}

The bimanual manipulator is provided with four skills implemented using:

\begin{enumerate}

    \item \emph{Motion planning}: A \texttt{Grasp} skill is based on the CBiRRT algorithm~\cite{CBiRRT2009} in the OpenRAVE simulator~\cite{OpenRAVE-Thesis}.

    \item \emph{Trajectory playback}: \texttt{Open}, \texttt{Scoop}, and \texttt{Spread} skills are implemented using relative end-effector trajectory playback modulated by impedance control.

\end{enumerate}

\paragraph*{Perception}

We use FoundationPose~\cite{foundationpose2024} for 6D object pose estimation, with poses initialized from CNOS~\cite{cnos2023} object segmentations.
These models require a textured mesh for each pose-estimated object.
We also provide a collision model of each object to the motion planner.


\subsubsection{Mobile Manipulation Setting}
\label{subsubsec:spot-experimental-setup}

\begin{figure*}[t!]
    \centering

    \subfigure[\texttt{OpenDrawer(c1)}]{ \includegraphics[width=0.27\linewidth]{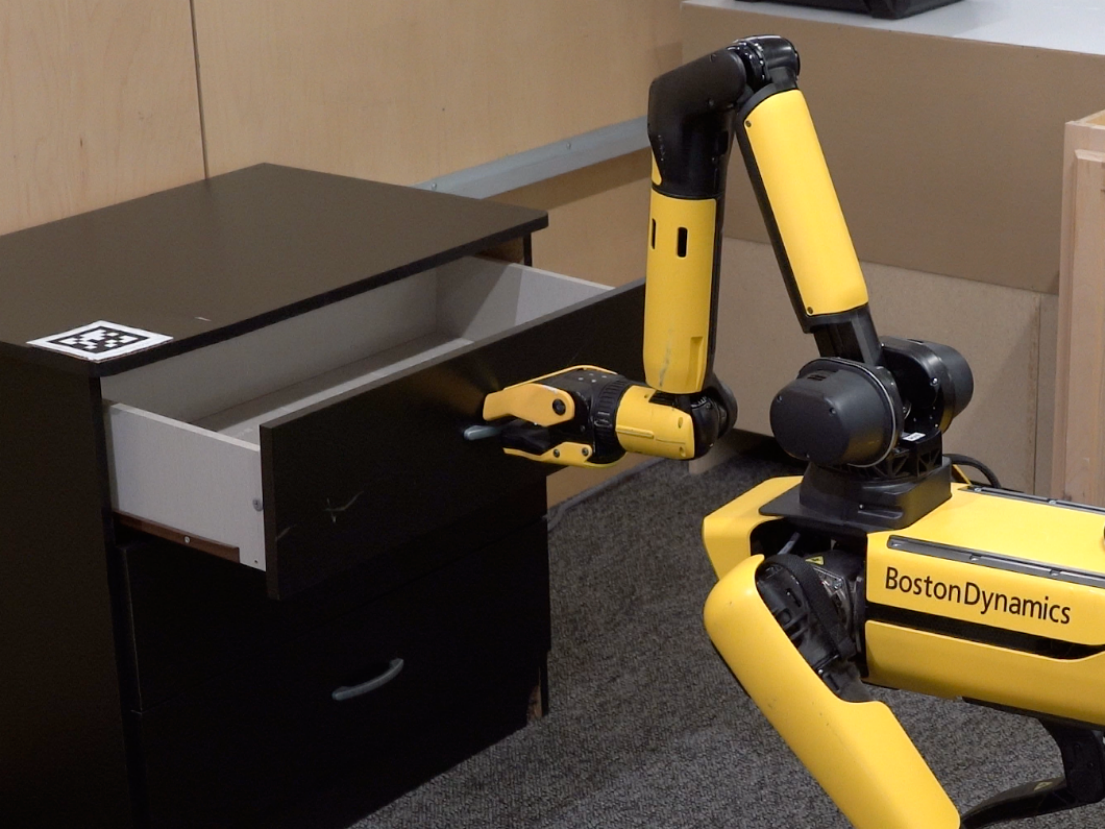}
    \label{fig:results-open-drawer}}
    \subfigure[\texttt{OpenDoor(d1)}]{\includegraphics[width=0.27\linewidth]{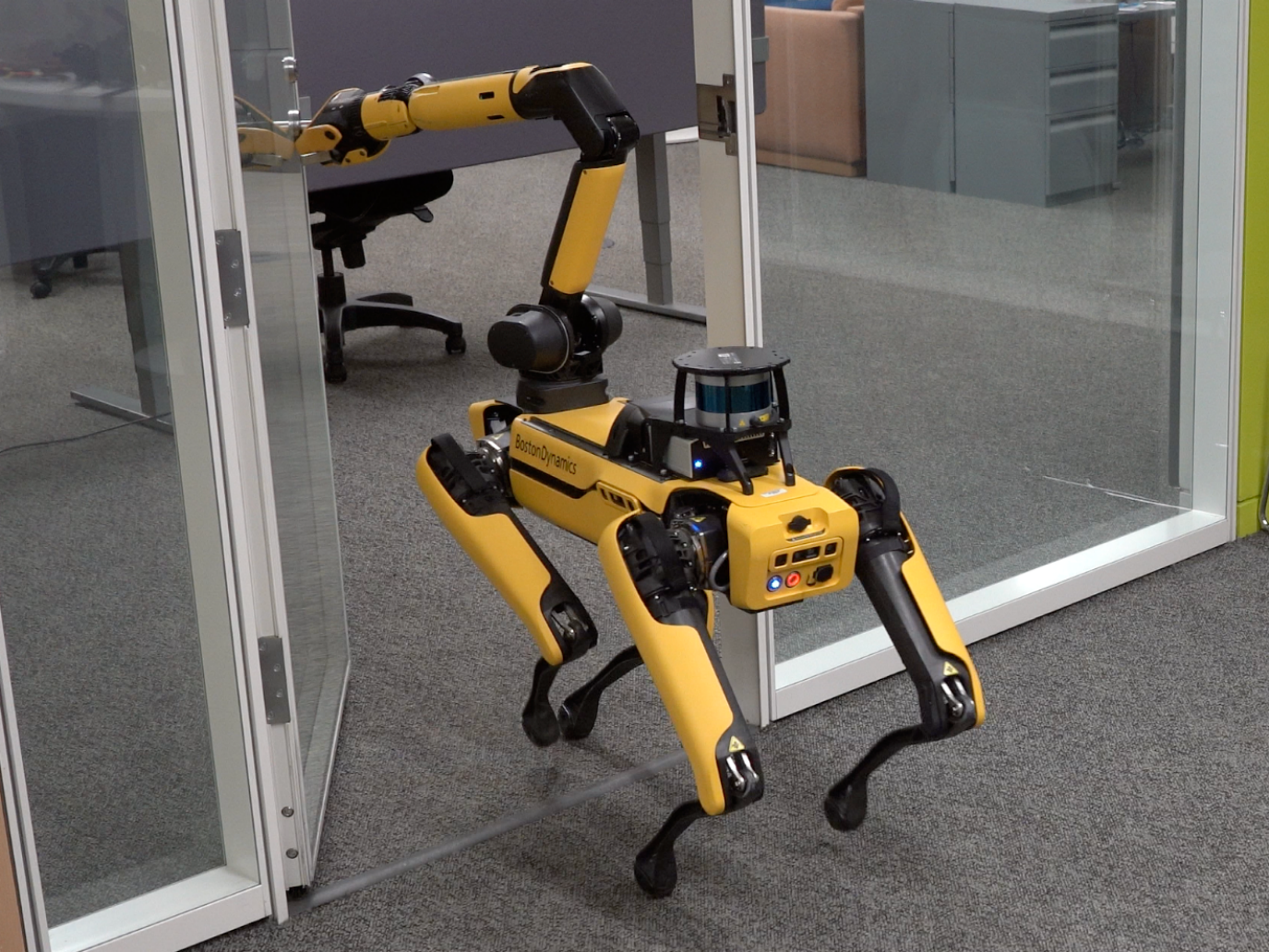}
    \label{fig:results-open-door}}
    \subfigure[\texttt{GoTo(f1)}]{ \includegraphics[width=0.27\linewidth]{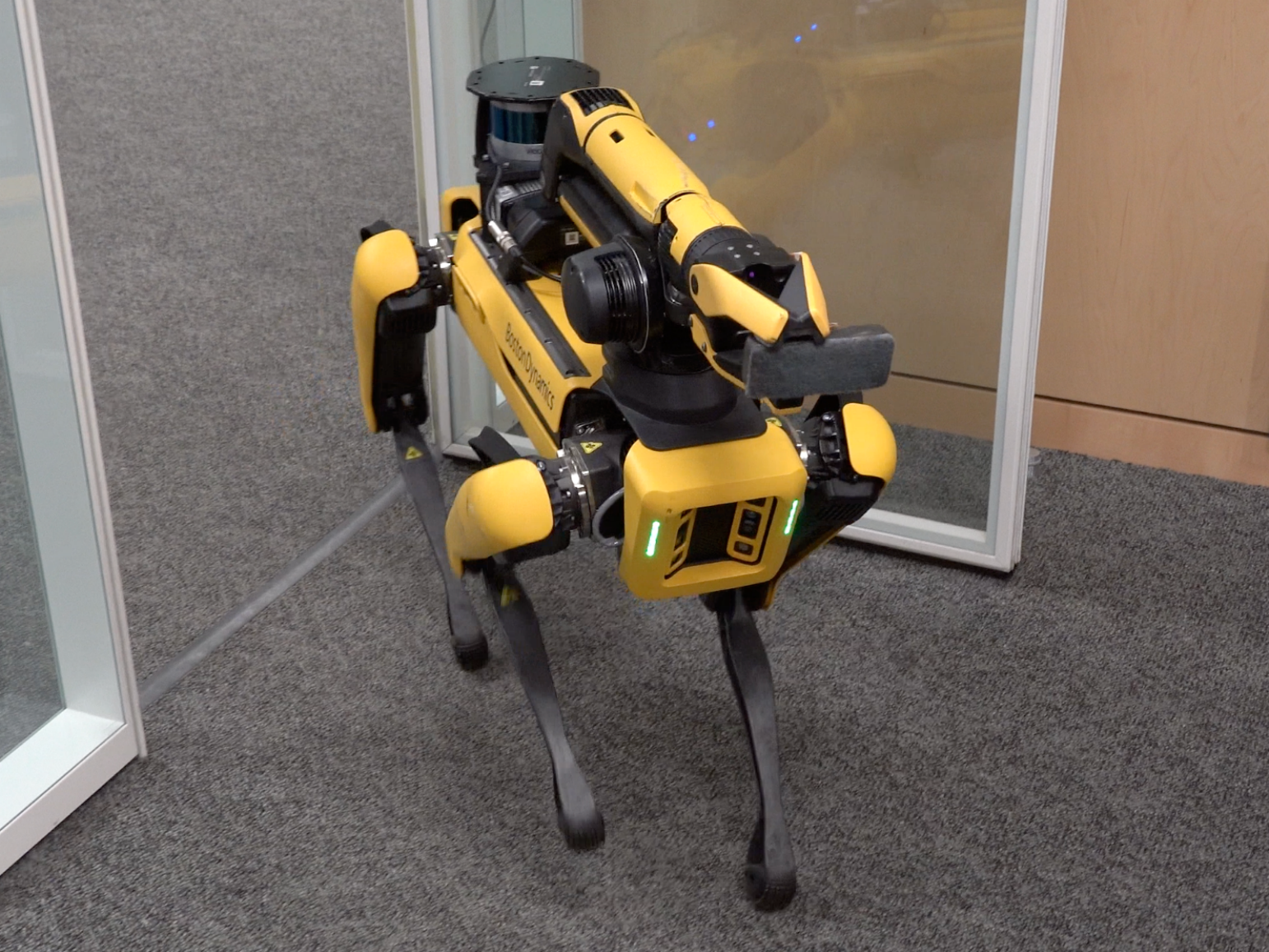}
    \label{fig:results-go-to}}

    \subfigure[\texttt{Place(e1,}~\texttt{f1)}]{%
    \includegraphics[width=0.27\linewidth]{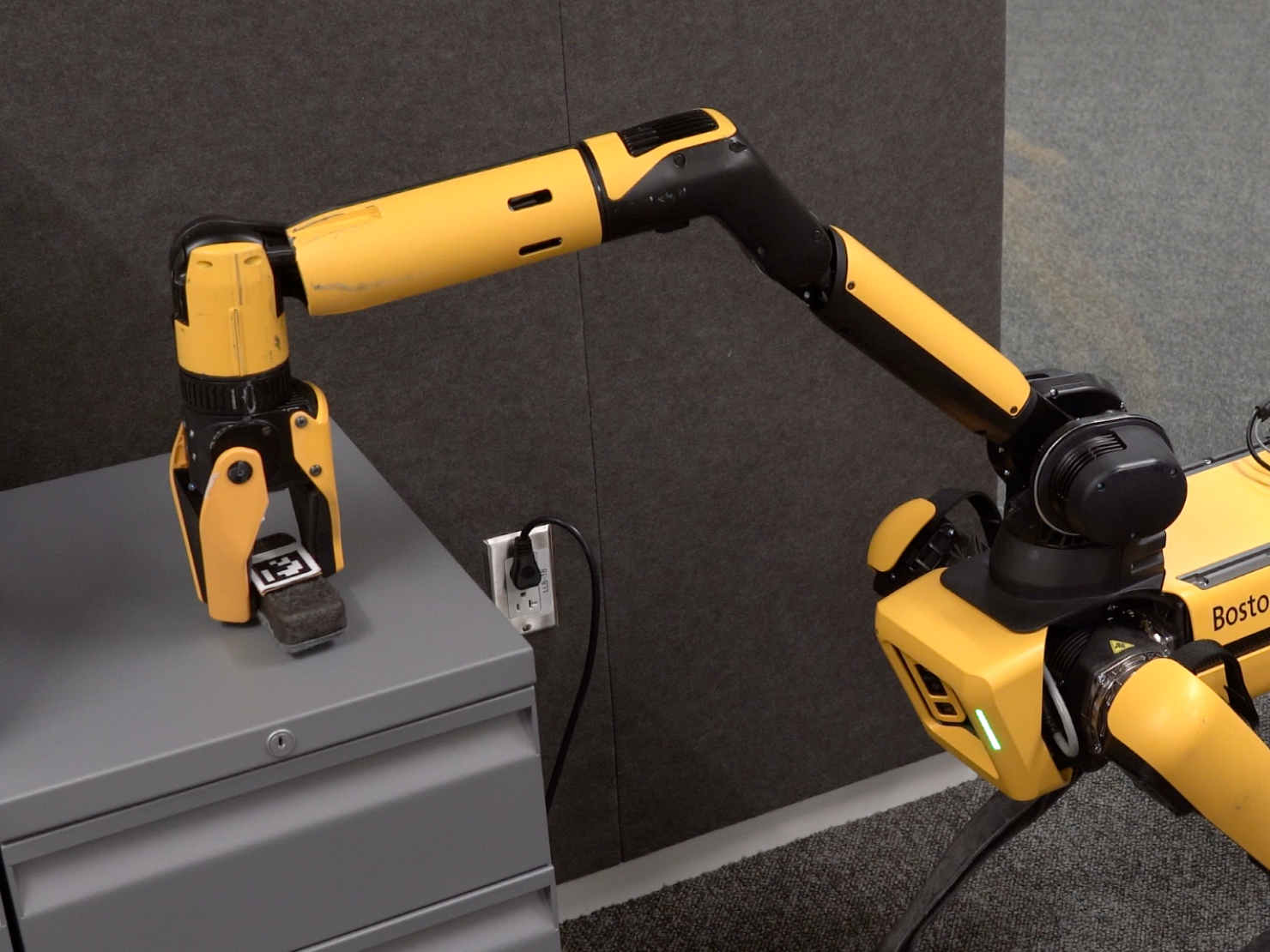}
    \label{fig:results-place}}
    \subfigure[\texttt{CloseDoor(d1)}]{%
    \includegraphics[width=0.27\linewidth]{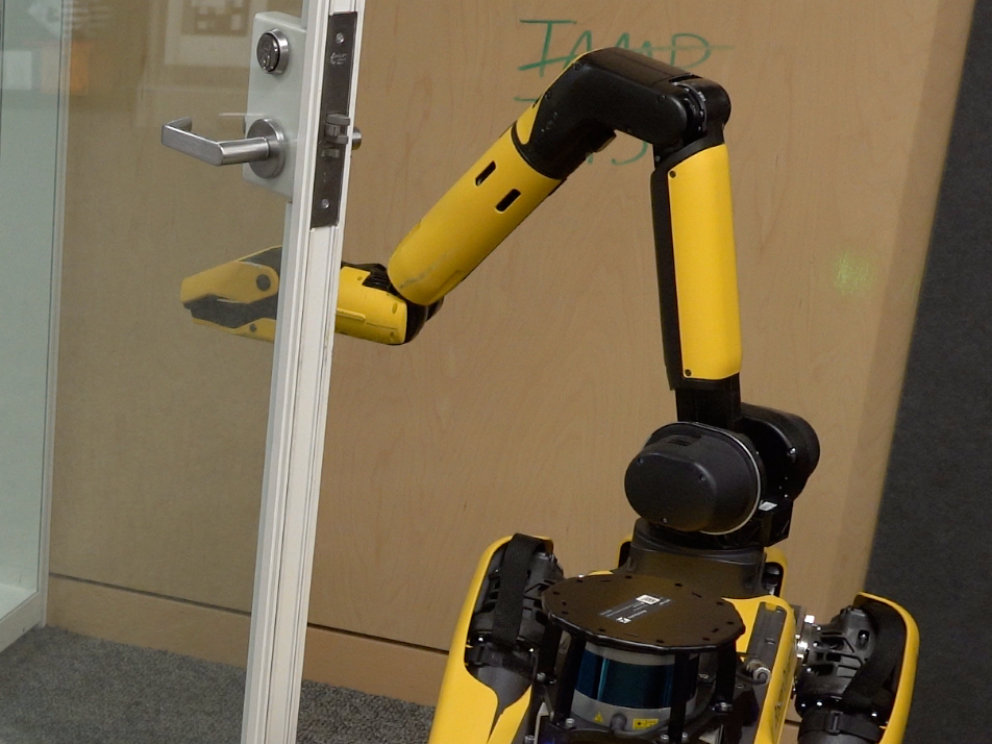}
    \label{fig:results-close-door}}
    \subfigure[\texttt{Erase(b1)}]{%
    \includegraphics[width=0.27\linewidth]{images/new-scili/Erase-4-3.pdf}
    \label{fig:results-erase}}
    
    \caption{We evaluate our method on a multi-room mobile manipulation task with non-monotonic structure and a heterogeneous skill inventory.
    The Spot robot opens a drawer using a trajectory playback skill \texttt{OpenDrawer(?drawer)} (Fig.~\ref{fig:results-open-drawer}), 
    opens a door using an off-the-shelf skill \texttt{OpenDoor(?door)} (Fig.~\ref{fig:results-open-door}),
    navigates and places the eraser using motion planning skills \texttt{GoTo(?location)} and \texttt{Place(?object,}~\texttt{?surface)} (Figs.~\ref{fig:results-go-to} and \ref{fig:results-place}),
    closes the door with a learned policy \texttt{CloseDoor(?door)} (Fig.~\ref{fig:results-close-door}),
    and erases the board using a force control skill \texttt{Erase(?board)} (Fig.~\ref{fig:results-erase}).}
    
    \label{fig:results-spot-execution}

\end{figure*}



\paragraph*{Robot Hardware}

In our mobile manipulation experiments, we use a Boston Dynamics Spot robot equipped with the Spot Arm manipulator and the Spot EAP sensor payload.
The robot's sensors include onboard RGB-D cameras, an RGB-D end-effector sensor, and a Velodyne VLP-16 LiDAR.
Due to onboard compute limitations, we run planning, learned policy control, and perception on an external machine with an NVIDIA GeForce RTX 4090 GPU.
%

\paragraph*{Robot Skill Inventory}

In our experiments, the Spot robot is equipped with an inventory of seven skills implemented using a diverse set of controllers:
\begin{enumerate}

    \item \emph{Motion planning}: We implement \texttt{Pick} and \texttt{Place} using MoveIt! \cite{moveit2014} and the Open Motion Planning Library \cite{ompl2012} for motion planning.
    
    \item \emph{Navigation}: The \texttt{GoTo} skill performs global planning with a grid-based A* planner and local control through the Boston Dynamics Spot SDK.

    \item \emph{Trajectory playback}: The \texttt{OpenDrawer} skill executes an object-relative end-effector trajectory with timed pauses and gradual gripper closure, exploiting gripper compliance to grasp the drawer handle reliably.

    \item \emph{Force control}: We implement an \texttt{Erase} skill using force control through the Boston Dynamics Spot SDK.

    \item \emph{Behavior cloning}: We implement a \texttt{CloseDoor} skill by training an Action Chunking Transformer~\citep{zhao2023learningfinegrainedbimanualmanipulation} policy on real-world demonstrations using LeRobot~\citep{cadene2024lerobot}.
    
    \item \emph{Off-the-shelf skill}: The \texttt{OpenDoor} skill detects the door handle using Gemini Robotics-ER 1.5~\citep{geminiroboticsteam2025geminirobotics15pushing} and then calls Spot's built-in door-opening functionality as a black-box skill.
    
\end{enumerate}

Across these seven skills, we highlight that only three rely primarily on motion planning (i.e., \texttt{Pick}, \texttt{Place}, and \texttt{GoTo}). Even for these skills, execution noise requires online pose estimation and, when needed, motion replanning.

\paragraph*{Perception Pipeline}

Before the experiment, the robot is teleoperated through the environment to construct an occupancy grid map.
During this phase, the robot also performs object pose estimation using AprilTag visual fiducial markers~\citep{AprilTag}.
The occupancy grid and estimated object poses are used to initialize the TASP system during planning.
We do not provide full kinematic models of articulated objects such as drawers or doors.
Instead, the system receives pre- and post-skill static models for each applicable skill, corresponding to the skill's kinematic envelope.



\subsection{Experimental Results}
\label{subsec:results}

We evaluate our approach using two real-world experiments designed to require long-horizon hierarchical planning that integrates motion planning and general-purpose skills.

\subsubsection{Bimanual Manipulation Experiment}
\label{subsubsec:dorfl-results}
In our bimanual manipulation experiment, the environment initially contains a kitchen knife \texttt{k0}, a closed jar of peanut butter \texttt{j0}, and a slice of bread \texttt{b0} (Fig.~\ref{fig:dorfl-grasp-pb}). The goal is to spread peanut butter on the bread.
In the solution plan produced by our planner, the robot first executes \texttt{Grasp(j0)} to pick up the jar, followed by \texttt{Open(j0)} to open it using both grippers (Figs.~\ref{fig:dorfl-grasp-pb}--\ref{fig:dorfl-open-pb}).
The robot then runs \texttt{Grasp(k0)} to pick up the knife using its available gripper (Fig.~\ref{fig:dorfl-grasp-knife}).
Finally, the robot executes \texttt{Scoop(k0,}~\texttt{j0)} to scoop peanut butter onto the knife, then spreads it onto the bread using \texttt{Spread(k0,}~\texttt{b0)} (Figs.~\ref{fig:dorfl-scoop} and \ref{fig:dorfl-spread}).

\subsubsection{Mobile Manipulation Experiment}
\label{subsubsec:spot-results}

Our mobile manipulation experiment tasks the Spot robot with erasing a wall-mounted writable board \texttt{b1} in a connecting room behind an initially closed door \texttt{d1}. When the door is open, it blocks access to the board, but when it is shut, Spot cannot enter the room.
That room also contains a filing cabinet \texttt{f1} near the board.
The robot begins in the same room as a closed chest of drawers \texttt{c1} containing a whiteboard eraser \texttt{e1}.

Given this task, the solution plan produced by our planner includes 14 high-level actions, seven of which are \texttt{GoTo(?location)} navigation skills alternating with manipulation skills.
For brevity, we therefore omit \texttt{GoTo} skills in the following description.

%
While in the initial room, Spot first uses \texttt{OpenDrawer(c1)} to open the chest's top drawer, then navigates to the door and executes \texttt{OpenDoor(d1)} (Figs.~\ref{fig:results-open-drawer} and \ref{fig:results-open-door}).
Spot then returns to the chest of drawers and runs \texttt{Pick(e1)} to pick the whiteboard eraser out of the opened drawer, as depicted in Figure~\ref{fig:1a-pick}.
With the eraser in hand, Spot heads into the other room and places the eraser onto the filing cabinet using \texttt{Place(e1,}~\texttt{f1)} (Figs.~\ref{fig:results-go-to} and \ref{fig:results-place}).
The robot then runs \texttt{CloseDoor(d1)} to close the door and make the board accessible (Fig.~\ref{fig:results-close-door}).
Finally, Spot picks the eraser back up using \texttt{Pick(e1)} and erases the text written on the board by executing \texttt{Erase(b1)} (Fig.~\ref{fig:results-erase}).
In both experiments, we observe that our approach can solve hybrid robot planning problems by integrating motion planning with general-purpose skills, including in scenarios with non-monotonic task progression.

\section{Related Work} \label{sec:related}

This work is most closely related to task and motion planning (TAMP), which combines discrete task-level reasoning with continuous motion planning to solve long-horizon problems~\citep{srivastava2014combined,toussaint2015logic,dantam2018incremental,shah2020anytime,garrett2021integrated}.
Recent work has extended TAMP-style methods beyond purely kinematic actions by incorporating learned skills~\citep{wang.garrett.ea2021,silver.athalye.ea2023,OptimisticRL2024}, closed-loop controllers~\citep{TAMPER2024,curtis2024partially}, and uncertainty-aware execution~\citep{curtis2024partially,garrett.paxton.ea2020}.
Accordingly, our contribution is not that skills can be combined with TAMP in general, but rather that our approach plans over a \emph{heterogeneous, pre-existing} inventory of robot skills while retaining geometric failure reasoning for non-downward-refinable actions~\citep{marthiAngelicSemanticsHighLevel2007}.

Prior approaches have incorporated learned skill models, learned feasibility or effect predictors, learned value functions, or controller libraries with explicitly modeled planning operators~\citep{wang.garrett.ea2021, silver.athalye.ea2023,liang.sharma.ea2022,STAP-Agia-et-al-2023,TAMPER2024,curtis2024partially}.
In contrast, we focus on planning with \emph{pre-existing, black-box} skills drawn from multiple implementation classes, including skills whose internal mechanics may be unavailable to the planner.
Our approach uses kinematic envelopes as a planner-facing geometric surrogate for such skills, allowing the planner to reason about where a skill can be executed and about its nominal geometric effects without modeling its internal controller dynamics.

Our methods are also related to prior work that combines task-level planning or TAMP with contact-rich, forceful, or reinforcement-learned behaviors~\citep{HolladayICRA2021, liang.cheng.eaCoRL2023, liu.deWinter.eaRAL2023, OptimisticRL2024, LEAGUE2023}.
We view these approaches as complementary: rather than emphasizing a single class of skill policy, we aim to support a mixed inventory of motion-planned, learned, force-controlled, trajectory-playback, and built-in robot skills within one planning framework.

\section{Conclusion and Future Work}
This paper formalizes a hybrid robot planning problem that requires the composition of general-purpose skills and motion planning.
We present a hierarchical approach to solving such planning problems, and demonstrate it on a bimanual manipulator and a mobile manipulator in challenging real-world scenarios.
%
%
Currently, we assume access to a predefined predicate vocabulary and per-skill kinematic envelopes. In future work, we intend to acquire these representations using world-model learning approaches that automatically learn initiation and termination sets, symbolic models, and kinematic envelopes for robot motor skills ~\citep{konidaris2018skills,shah2024reals}. 

\section*{Acknowledgment}
This work was supported in part by ONR REPRISM
MURI N00014-24-1-2603, ONR grant 00014-22-1-2592, and the
Robotics and AI Institute (RAI).

\bibliographystyle{IEEEtranSN}
\bibliography{ref}

\end{document}